\pdfoutput=1
\documentclass[11pt]{article}

\usepackage[final]{coling}

\usepackage{times}
\usepackage{latexsym}

\usepackage[T1]{fontenc}
\usepackage{CJKutf8}
\usepackage[whole]{bxcjkjatype}

\usepackage[utf8]{inputenc}

\usepackage{microtype}

\usepackage{inconsolata}

\usepackage{graphicx}

\usepackage{booktabs}
\usepackage{multirow}
\usepackage{makecell}
\usepackage{xcolor}
\usepackage{listings}
\usepackage{CJKutf8}
\newcommand{\ja}[1]{\begin{CJK}{UTF8}{ipxm}#1\end{CJK}}
\newcommand{\zh}[1]{\begin{CJK*}{UTF8}{gbsn}#1\end{CJK*}}

\title{MQM-Chat: Multidimensional Quality Metrics for Chat Translation}
\author{Yunmeng Li$^{1}$
Jun Suzuki$^{1,2}$ 
Makoto Morishita$^{3,1}$ 
Kaori Abe$^{4,1,2}$
Kentaro Inui$^{5,1,2}$ \\
${}^{1}$Tohoku University
${}^{2}$RIKEN \\
${}^{3}$Future Corporation
${}^{4}$Machine Learning Solutions Inc.
${}^{5}$MBZUAI\\
\texttt{li.yunmeng.r1@dc.tohoku.ac.jp}
}
\begin{document}
\maketitle
%%%%%%%%%%%%%%%%%%%%%%%%%%%%%%%%%%%%%%%%%%%%%%%%%%%%%%%%%%%%%%%%%%%%%%%%%%%%%%%%%%%%%%%%%%%%%%%%%%%%
\begin{abstract}

The complexities of chats, such as the stylized contents specific to source segments and dialogue consistency, pose significant challenges for machine translation.
Recognizing the need for a precise evaluation metric to address the issues associated with chat translation, this study introduces Multidimensional Quality Metrics for Chat Translation (MQM-Chat), which encompasses seven error types, including three specifically designed for chat translations: ambiguity and disambiguation, buzzword or loanword issues, and dialogue inconsistency. 
In this study, human annotations were applied to the translations of chat data generated by five translation models.
Based on the error distribution of MQM-Chat and the performance of relabeling errors into chat-specific types, we concluded that MQM-Chat effectively classified the errors while highlighting chat-specific issues explicitly.
The results demonstrate that MQM-Chat can qualify both the lexical accuracy and semantical accuracy of translation models in chat translation tasks.

\end{abstract}
%%%%%%%%%%%%%%%%%%%%%%%%%%%%%%%%%%%%%%%%%%%%%%%%%%%%%%%%%%%%%%%%%%%%%%%%%%%%%%%%%%%%%%%%%%%%%%%%%%%%
\section{Introduction}

Recent developments in neural machine translation (NMT)~\cite{bahdanau2014neural, patil2014use, medvedev2016google, gehring2017convolutional} have demonstrated notable improvements in the performance of machine translation systems, especially in tasks involving the translation of formatted documents such as news articles and academic papers~\cite{maruf-haffari-2018-document, barrault-etal-2019-findings, barrault-etal-2020-findings, nakazawa-etal-2019-overview, ma-etal-2020-simple}.
However, current methods continue to face considerable challenges when translating chats~\cite{tiedemann-scherrer-2017-neural, maruf-etal-2018-contextual, farajian-etal-2020-findings} due to their high degrees of ambiguity and stylized contents, including sentiments, personalities, and cultural nuances~\cite{uthus2013multiparticipant, laubli-etal-2018-machine, toral-etal-2018-attaining, farajian-etal-2020-findings} %baldwin2013noisy, eisenstein2013bad, xu2014empirical.
To enhance the performance of chat translation, it is important to understand the qualities and limitations of existing translation models in handling chats.
% \usepackage{graphicx}
\begin{table*}[t]
\centering
\small
%\resizebox{\textwidth}{!}{%
% \usepackage{booktabs}
\begin{tabular}{@{}lll@{}}
\toprule
\textit{\textbf{Source (zh, ja)}} & \textbf{Possible Good Translation (en)} & \textbf{Bad Translation (en) by MT Model} \\ \midrule

\multicolumn{3}{c}{\textit{Ambiguity and Disambiguation}} \\ \midrule
\textcolor{blue}{\zh{队啊}}\zh{！你应该试试！} & \textcolor{blue}{Yaas}! You should try! & \textcolor{red}{Team ah}! You should try! \\ 
\textcolor{blue}{\ja{知ってｒ？}}\ja{昨日、ヘレンとあったよ} & \textcolor{blue}{u know waht}, I saw Helen yesterday & \textcolor{red}{You know what}, I saw Helen yesterday \\ \midrule

\multicolumn{3}{c}{\textit{Buzzword or Loanword Issues}} \\ \midrule
\textcolor{blue}{\zh{鼠的}}\zh{，真的累死了} & \textcolor{blue}{Yaap}, I'm really tired & \textcolor{red}{Rat's}, I'm really exhausted \\ 
\textcolor{blue}{\ja{草}wwwwww} & \textcolor{blue}{lol} & \textcolor{red}{grass} \\ \midrule

\multicolumn{3}{c}{\textit{Dialogue Inconsistency}} \\ \midrule
\zh{你觉得我能赢吗？} & Do you think I can win? & Do you think I can win? \\
- \zh{没事儿，肯定能赢啊！} & - It's okay. I'm sure \textcolor{blue}{you}'ll win! & - It's okay. I'm sure \textcolor{red}{they}'ll win! \\ \midrule
\ja{まどかは昨日買い物に行ったよ} & Madoka went shopping yesterday. & Madoka went shopping yesterday. \\
- \ja{行った？聞いてないよ！} & - \textcolor{blue}{She} went? I didn't hear about it! & - \textcolor{red}{You} went? I didn't hear about it! \\ \bottomrule
\end{tabular}
%}
%\caption{Examples of \textit{Ambiguity and Disambiguation}, \textit{Buzzword or Loanword Issues} and \textit{Dialogue Inconsistency} errors. Translations in blue are possibly expected, and translations in red are bad.}
\caption{Examples of MQM-Chat's chat-specific errors, including \textit{Ambiguity and Disambiguation}, \textit{Buzzword or Loanword Issues}, and \textit{Dialogue Inconsistency}.}
\label{tab:error_type_examples}
\end{table*}
% --------------------------------------------- %
Traditional automatic evaluation metrics, such as BLEU~\cite{papineni-etal-2002-bleu} and COMET~\cite{rei-etal-2022-comet, rei-etal-2022-cometkiwi}, primarily focus on accuracy; however, they fail to capture the meanings and the stylized contents, especially when evaluating chats.
For example, as shown in Table~\ref{tab:error_type_examples}, translation models generate errors because of source-side typographical errors (typos), Internet slang, and the omission of subjects in the flow of chats.
When evaluating the chat translation quality, we need to focus on nuances, stylized content specific to the source segments, and dialogue consistency in addition to grammatical and lexical accuracy.
Thus, a refined error categorization framework that can assess semantic accuracy while preserving the chat-specific nuances and content is better suited for chat translation tasks~\cite{gehman-etal-2020-realtoxicityprompts}.

In this paper, we propose the Multidimensional Quality Metrics for Chat Translation (MQM-Chat) to address the challenges of evaluating chat translations.
Building on the existing Multidimensional Quality Metrics (MQM) framework\footnote{\url{https://themqm.org/}}~\cite{burchardt-2013-multidimensional, mariana2014multidimensional, freitag-etal-2021-experts}, MQM-Chat incorporates seven error types: \textit{Mistranslation}, \textit{Omission or Addition}, \textit{Terminology or Proper Noun Errors}, \textit{Unnatural Style}, \textit{Ambiguity and Disambiguation}, \textit{Buzzword or Loanword Issues}, and \textit{Dialogue Inconsistency}.
The latter three were specifically designed to handle the chat nuances.

MQM-Chat was applied to evaluate the chat translation capabilities of five models: the large language models (LLMs) GPT-4~\cite{achiam2023gpt} and LLaMA3~\cite{llama3}, the commercial translation model DeepL\footnote{\url{https://www.deepl.com/translator}}, the multilingual model by Facebook at WMT21~\cite{tran-etal-2021-facebook}, and the bilingual model by team SKIM at WMT23~\cite{kudo-etal-2023-skim}.
We translated Chinese (zh) and Japanese (ja) chat data into English (en) using these models and assigned human annotations.
MQM-Chat helped highlight the issues and qualified the strengths and weaknesses of the five models crossing language pairs.

% \usepackage{booktabs}
% \usepackage{multirow}
% \usepackage{makecell}
% \usepackage{graphicx}
\begin{table*}[ht]
\centering
\small
\resizebox{\textwidth}{!}{%
\begin{tabular}{llllcl}
\toprule
\textbf{} &
  \textbf{Chat Domain} &
  \makecell{\textbf{Human}\\
  \textbf{Evaluation Method}} &
  \textbf{Evaluation Focus} &
  \makecell{\textbf{Fine-grained}\\
  \textbf{Analysis}} &
  \textbf{Language Pairs} \\
  \midrule
  
\makecell[l]{\textbf{WMT 2020} \\
\textbf{Chat Translation}} &
  Custom Service &
  \makecell[l]{Segment Rating\\
  + Document Context} &
  Pronoun (\textit{it}). &
  $\triangle$ &
  en$\Leftrightarrow$de \\
  \midrule
  
\makecell[l]{\textbf{WMT 2022} \\
\textbf{Chat Translation}} &
  Custom Service &
  Adapted MQM* &
  \makecell[l]{Accuracy,\\
  Linguistic Conventions, \\
  Terminology, ...\\
  %style,\\
  %locale convention, \\
  %audience appropriateness, \\
  %design and markup, \\
  MT Hallucination, \\
  Source Issue.} &
  $\triangle$ &
  \makecell[l]{en$\Leftrightarrow$de,\\
  en$\Leftrightarrow$fr,\\
  en$\Leftrightarrow$pt\_br} \\
  \midrule
  
\textbf{CPCC} &
  \makecell[l]{Custom Service,\\TV series} &
  Customized &
  \makecell[l]{Preference, Coherence,\\
  Consistency, Fluency.} &
  $\bigcirc$ &
  \makecell[l]{en$\Leftrightarrow$de,\\
  en$\Leftrightarrow$zh} \\
  \midrule
  
\textbf{CSA-NCT} &
  \makecell[l]{Custom Service,\\TV series} &
  Customized &
  \makecell[l]{Coherence, Speaker, \\
  Fluency.} &
  $\bigcirc$ &
  \makecell[l]{en$\Leftrightarrow$de,\\
  en$\Leftrightarrow$zh} \\
  \midrule
  
\textbf{SML} &
  \makecell[l]{Custom Service,\\TV series} &
  Question-based &
  Coherence, Fluency. &
  $\bigcirc$ &
  \makecell[l]{en$\Leftrightarrow$de,\\
  en$\Leftrightarrow$zh} \\
  \midrule
  
%\textbf{MSCTD} &
%  Building Dataset &
%  Dialogue from Movies (OpenViDal) with Sentiments &
%  \multicolumn{1}{c}{/} &
%  / &
%  \multicolumn{1}{c}{/} &
%  en$\Leftrightarrow$ de, en$\Leftrightarrow$ zh \\
\makecell[l]{\textbf{MQM-Chat}\\
\textbf{Annotation}} &
  \makecell[l]{\textbf{Various}\\
  (news, sports,\\
  hobbies, daily life,\\
  social media, etc.)} &
  MQM-Chat &
  \makecell[l]{
  Source Issue$\rightarrow$\textbf{Disambiguation},\\
  Consistency$\rightarrow$\textbf{Dialogue Consistency},\\
  Speaker$\rightarrow$\textbf{Stylized Contents},\\
  \textbf{Cultural Contents},\\
  \textbf{Buzzwords and Loanwords}.\\} &
  $\bigcirc$ &
  \makecell[l]{zh$\Rightarrow$en\\
  \textbf{ja}$\Rightarrow$en} \\ 
  \bottomrule
\end{tabular}%
}
\caption{Comparison of previous studies across several dimensions: data domain, human evaluation method, evaluation focus, granularity of results, and language pairs studied.}%WMT2020 and WMT2022 analyses are considered less detailed due to the Segment Rating + Document Context method and lack of fine-grained explanations on terminal nodes.}
\label{tab:research_comparison}
\end{table*}
% --------------------------------------------- %

To verify the effectiveness of MQM-Chat, we compare it with the standard MQM framework. 
Standard MQM human annotations were assigned to sampled data, and the differences between the two approaches were analyzed.
Our findings demonstrate that MQM-Chat provides fine-grained classification, recognizing a significant portion of errors in standard MQM into other labels, with approximately 30\% of these as chat-specific issues such as \textit{Ambiguity and Disambiguation}, \textit{Buzzword or Loanword Issues}, and \textit{Dialogue Inconsistency}.

We have attempted to implement automatic annotation using MQM-Chat with few-shot learning.
The results indicated that the auto annotations obtained by MQM-Chat agree with the overall system performance annotated by human annotators but are not as accurate as human annotations.

In summary, this study contributes to the chat translation field by proposing the MQM-Chat evaluation metric.
Five state-of-the-art translation models were evaluated with MQM-Chat when handling chat content.
The experiments helped construct annotated zh$\Rightarrow$en and ja$\Rightarrow$en chat translation data.
The contributions of this study are expected to enhance the understanding of chat translation and provide valuable resources for future advancements in the field.

%%%%%%%%%%%%%%%%%%%%%%%%%%%%%%%%%%%%%%%%%%%%%%%%%%%%%%%%%%%%%%%%%%%%%%%%%%%%%%%%%%%%%%%%%%%%%%%%%%%%
\section{Related Work}

%%%%%%%%%%%%%%%%%%%%%%%%%%%%%%
\subsection{Translation Evaluation Metrics}

Traditional metrics such as BLEU~\cite{papineni-etal-2002-bleu}, which utilizes n-grams, 
and METEOR~\cite{banerjee-lavie-2005-meteor}, which considers an alignment using unigrams, 
rely on the textual similarity between the model's output and reference texts to produce evaluation scores.
Additionally, BERTScore~\cite{zhang2020bertscoreevaluatingtextgeneration}, which measures semantic similarity, and generation-based metrics like Prism~\cite{easy_doc_mt}, rely on the textual similarity between the model’s output and reference texts to produce evaluation scores.
Recent metrics, such as BERTScore~\cite{zhang2020bertscoreevaluatingtextgeneration}, COMET~\cite{rei-etal-2020-comet} and BLEURT~\cite{sellam-etal-2020-bleurt}, measure the semantic similarity using language models, and they employ neural networks to enhance score generation. 
These metrics focus more on semantic understanding by aligning their assessments with human evaluations.

The Multidimensional Quality Metrics (MQM) framework~\cite{burchardt-2013-multidimensional, mariana2014multidimensional, freitag-etal-2021-experts} is a detailed and flexible approach to further refine the task of evaluating translation quality. 
It assesses word-level errors, semantic accuracy, stylistic nuances, and cultural appropriateness.
The MQM Core~\footnote{\url{https://themqm.org/the-mqm-typology/}} includes 39 distinct error types, and the MQM Full~\footnote{\url{https://themqm.org/the-mqm-full-typology/}} includes even more, offering a comprehensive translation quality assessment.
In addition to error types, the MQM framework allows annotators to tag each error with a severity to indicate its impact on the translation.
Recent developments in automatic MQM evaluation include xCOMET~\cite{guerreiro2023xcomettransparentmachinetranslation}, AutoMQM~\cite{fernandes-etal-2023-devil}, and GEMBA-MQM~\cite{kocmi-federmann-2023-gemba}.
AutoMQM and GEMBA-MQM leverage LLMs to automatically detect and label the errors with MQM error types and severity.
Compared to existing metrics, we defined a benchmark that provides a multidimensional evaluation with error types specific to chat translation tasks.

%%%%%%%%%%%%%%%%%%%%%%%%%%%%%%
\subsection{Chat Translation Tasks}

%While formal documents follow standardized structures, 
Chats frequently include slang, idiomatic expressions, and personalized styles, increasing the translation task's complexity~\cite{baldwin-etal-2013-noisy, eisenstein-2013-bad}.
While high accuracy is important when translating chats, preserving the nuances and special contents is sometimes even more crucial~\cite{hovy2015demographic, salganik2020digital}.

The first workshop that focused on chat translation was the Fifth Conference on Machine Translation (WMT20)~\cite{barrault-etal-2020-findings, farajian-etal-2020-findings}, which laid the groundwork in this domain.
This was followed by the Seventh Conference on Machine Translation (WMT22)~\cite{kocmi-etal-2022-findings, farinha-etal-2022-findings} and continued with the Ninth Conference on Machine Translation (WMT24)~\cite{mohammed-etal-2024-findings}.
The WMT shared tasks have primarily focused on customer service chats, which are relatively structured and standardized.
The emphasis has been on evaluating the overall performance of chat translation models with a strong focus on syntax accuracy.
The WMT2022 shared task began to address chat-specific issues, and Liang's team, as a continuation of WMT2020, presented improved chat translation models, highlighting the importance of coherence, fluency, and speaker personalities~\cite{liang-etal-2021-modeling, liang-etal-2021-towards, liang-etal-2022-scheduled}.

Gradually, WMT and derivative studies have recognized the importance of source content issues and preserving the speaker's style in chat translations.
Thus, MQM was adapted in the WMT2022 shared task; however, it was too broad with 31 error types, most of which focused on accuracy and relatively superficial analyses.
%Liang's studies focused on personality and sentiment but did not consider source issues.
Previous studies have utilized binary classification for chat translation with a specific focus on coherence~\cite{li-etal-2022-chat, li-etal-2023-investigation}, which did not capture the complexity of chat translations effectively.

With these foundations, we have refined the evaluation process by differentiating the source issues in chat translations into ambiguity issues and cultural nuances issues such as buzzwords, and by emphasizing the importance of dialogue consistency.
Additionally, we have de-emphasized grammatical accuracy, which is not always the highest priority in everyday conversations.
To make MQM-Chat broadly applicable to general chats, we chose data that covers various topics, including news, sports, hobbies, daily life, and social media.
Furthermore, we have included Japanese data, a language that has not been studied extensively in chat translation tasks.
A comparison between our research and previous studies is presented in Table~\ref{tab:research_comparison}.

%%%%%%%%%%%%%%%%%%%%%%%%%%%%%%
%\paragraph{Translating with LLMs}

%Several studies have demonstrated that GPT performs well in translation tasks~\cite{hendy2023good, zhang2023prompting}, particularly in scenarios involving literary translation~\cite{thai-etal-2022-exploring, karpinska-iyyer-2023-large}.
%These studies suggest that LLM translations might be favored over traditional neural machine translation (NMT) models when the input domain is likely to contain noisy, ill-formed sentences.
%Despite these promising findings, no dedicated research specifically addresses chat translation using LLMs.
%This gap highlights the need for focused studies on applying LLMs to the unique challenges of chat translation.

%%%%%%%%%%%%%%%%%%%%%%%%%%%%%%%%%%%%%%%%%%%%%%%%%%%%%%%%%%%%%%%%%%%%%%%%%%%%%%%%%%%%%%%%%%%%%%%%%%%%
\section{Multidimensional Quality Metrics for Chat Translation (MQM-Chat)}

In this study, we define high-quality chat translation as maintaining accuracy while simultaneously capturing and conveying the speaker's personality, styles, and cultural nuances effectively.
We refined the MQM framework and introduced customized categories that are specific to chat translation tasks.

%%%%%%%%%%%%%%%%%%%%%%%%%%%%%%%%%%%%%%%%%%%%%%%%%%
\subsection{Error Types}

% --------------------------------------------- %
\begin{figure*}[ht]
\small
\centering
\includegraphics[width=12cm]{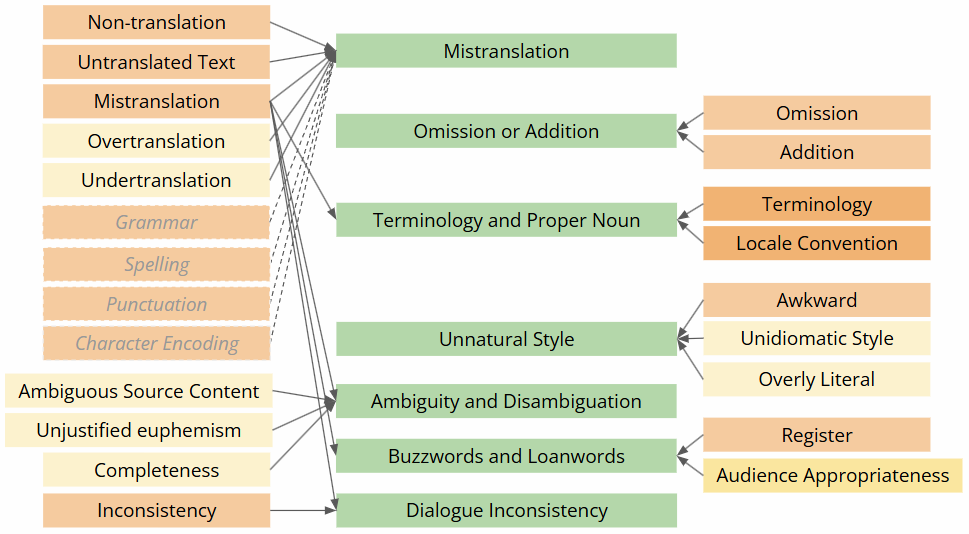}
\caption{Mapping of error types in MQM-Chat (green) and MQM Core (orange) and used MQM Full (yellow). 
Blocks with deeper colors (\textit{Terminology}, \textit{Locale Convention}, and \textit{Audience Appropriateness}) suggest that corresponding sub-categories are included and merged into MQM-Chat. Blocks with gray text (\textit{Grammar}, \textit{Spelling}, \textit{Punctuation}, \textit{Character Encoding}) are errors that are only marked if they totally interrupt the translation substantially. Note that the relationship between the mapping blocks is not simply an inclusion relationship because MQM-Chat error types cover broader issues in chat translation.}
\label{fig:error_types_map}
\end{figure*}

% --------------------------------------------- %

As mentioned previously, MQM-Chat focuses on seven error types: \textit{Mistranslation}, \textit{Omission or Addition}, \textit{Terminology or Proper Noun Issues}, \textit{Unnatural Style}, \textit{Ambiguity and Disambiguation}, \textit{Buzzwords or Loanwords Issues}, and \textit{Dialogue Inconsistency}. 
The latter three error types are customized typologies for chat translation.
The errors are evaluated with three levels of severity to provide a sufficiently detailed and accurate assessment.
The mapping of error types between MQM-Chat and standard MQM is shown in Figure~\ref{fig:error_types_map}.
Note that the relationship between the mapping blocks is not just an inclusion relationship because error types in MQM-Chat cover broader issues in chat translation tasks with specific descriptions and examples.

%%%%%%%%%%%%%%%%%%%%%%%%%%%%%%
\paragraph{Mistranslation}

\textit{Mistranslation} refers to fundamental inaccuracies in the translation process, %resulting in deviations from the intended meaning of the source text.
including untranslated source segments, incorrect lexical choice or grammar that distorts the meaning, as well as undertranslation and overtranslation.
%insufficient detail or generality compared to the source (under-translation), and excessive specificity compared to the source (over-translation).
These errors are critical because they directly impact the comprehensibility and accuracy of the translation.

%%%%%%%%%%%%%%%%%%%%%%%%%%%%%%
\paragraph{Omission or Addition}

%Omission or Addition errors involve discrepancies where content present in the source text is missing in the translation (omission), or content not present in the source text is included in the translation (addition).
Missing source content (omission) or additional content not present in the source (addition) are considered to be \textit{Omission or Addition} errors.
Such errors can significantly mistake the intended message and disrupt the coherence of the translated text, which can result in misunderstandings.

%%%%%%%%%%%%%%%%%%%%%%%%%%%%%%
\paragraph{Terminology and Proper Noun Issues}

\textit{Terminology and Proper Noun Issues} are related to inaccuracies when translating specialized vocabulary, inherent terms, and proper nouns from the source text. 
Misinterpretations in this category can undermine the reliability of the translation, especially in professional and academic contexts.
Note that this category does not include Internet terms, popular terms, newly created words, memes, and foreign words.

%%%%%%%%%%%%%%%%%%%%%%%%%%%%%%
\paragraph{Unnatural Style}

\textit{Unnatural Style} refers to translations that are grammatically correct but unnatural in the target language.
%These errors affect the readability and acceptability of the translation, making it appear awkward or stilted to native speakers.

%%%%%%%%%%%%%%%%%%%%%%%%%%%%%%
\paragraph{Ambiguity and Disambiguation}

The goal of this study is to retain the speaker-specific stylized contents and accurately translate them into their corresponding errors in the target language. 
\textit{Ambiguity and Disambiguation} errors occur when the ambiguities or errors in the source text, such as typographical errors, omissions, unclear abbreviations, and erroneous punctuation, are not faithfully reflected in the translation.
For example, the typos \ja{知ってr} and \zh{队啊} shown in Table~\ref{tab:error_type_examples} are considered to be ambiguity errors.
Deviations from this principle are considered errors, which highlights the need to translate the speaker-specific stylized content into corresponding errors in the target language.
This error category emphasizes the importance of maintaining the authenticity of the source text, including its imperfections.

%%%%%%%%%%%%%%%%%%%%%%%%%%%%%%
\paragraph{Buzzword or Loanword Issues}

\textit{Buzzword or Loanword Issues} occur when such terms are not translated accurately according to their usage in both the source and target languages.
This includes the incorrect translation of popular sayings, newly created words, Internet slang, and memes.
For example, in Table~\ref{tab:error_type_examples}, the Japanese meme \ja{草} (representing laughter) and the Chinese Internet slang \zh{鼠的} (meaning yes) are frequently mistranslated by existing translation models, capturing only the superficial aspects of their usage. 
This results in errors that obscure the source text's intended meaning and cultural nuance of the source text.
Thus, if there is no corresponding term in the target language, the pronunciation should be retained and written in the target language.

%%%%%%%%%%%%%%%%%%%%%%%%%%%%%%
\paragraph{Dialogue Inconsistency}

\textit{Dialogue Inconsistency} occurs when translations fail to maintain consistency based on context, particularly when the speakers change in the chat.
This can include inappropriate handling of demonstrative pronouns, personal references, or definite articles.
For example, in Japanese and Chinese, the subject is frequently omitted when it has already appeared in the preceding context. 
As shown in Table~\ref{tab:error_type_examples}, the subjects \zh{你} (you) in the Chinese source and \ja{まどか} (Madoka, she) in the Japanese source are omitted, which results in errors.
Maintaining sufficient consistency in dialogue is crucial to ensure coherence and avoid confusing the reader.

%%%%%%%%%%%%%%%%%%%%%%%%%%%%%%%%%%%%%%%%%%%%%%%%%%
%\subsection{Error Severity Levels}

%We provided three levels of severity for each error to evaluate the translations comprehensively: \textbf{major} for errors that significantly impact the understandability of the content; \textbf{minor} for errors that do not impact the overall understandability but detract from the quality; \textbf{neutral} for errors requiring additional revision but do not pose significant risks.
%Severity penalty multipliers are 5 for major, 3 for minor, and 1 for neutral.

%%%%%%%%%%%%%%%%%%%%%%%%%%%%%%%%%%%%%%%%%%%%%%%%%%%%%%%%%%%%%%%%%%%%%%%%%%%%%%%%%%%%%%%%%%%%%%%%%%%%
\section{Experiments}

Experiments were conducted to evaluate the effectiveness of MQM-Chat by translating chats from Japanese (ja) and Chinese (zh) into English (en) and assigning the MQM-Chat human annotations, MQM-Chat automatic annotations, and standard MQM automatic annotations.

%%%%%%%%%%%%%%%%%%%%%%%%%%%%%%%%%%%%%%%%%%%%%%%%%%
\subsection{Datasets}

In the experiments, 200 short chats were selected from the Open 2ch Dialogue Corpus~\cite{open2chdlc2019en} for the ja$\Rightarrow$en translations, which feature ambiguous content and popular sayings from Japan's online community 2channel.
Sensitive content was excluded to avoid offensive data.
Similarly, we selected 200 short chats from the LCCC-base dataset~\cite{wang2020chinese} for the zh$\Rightarrow$en translations.
To provide an effective comparison and a broader range of chat contents, we included 100 long chats from BPersona-chat~\cite{sugiyama-etal-2021-empirical, li-etal-2022-chat} for ja$\Rightarrow$en and 100 long chats from NaturalConv~\cite{wang2021naturalconv} for zh$\Rightarrow$en.
Data statistics are listed in Table~\ref{tab:data_stat} in Appendix~\ref{app:datasets}.

%%%%%%%%%%%%%%%%%%%%%%%%%%%%%%%%%%%%%%%%%%%%%%%%%%
\subsection{Translation Models}

Several translation models were considered in the experiments, including sentence-to-sentence transformers-based models~\cite{vaswani2017attention}, LLMs, and commercialized systems to generate the translation data.
Specifically, they are GPT-4, LLaMA3 (70B-Instruct), DeepL, Facebook@WMT21 for zh$\Rightarrow$en and SKIM@WMT23 for ja$\Rightarrow$en.
GPT-4 and LLaMA3 were used in zero-shot learning configurations~\cite{romera2015embarrassingly, wang2019survey} with prompts designed on methodologies proposed by \citet{hendy2023good} and other recent studies~\cite{farinhas-etal-2023-empirical, peng-etal-2023-towards}.
Detailed prompts and model parameters are listed in Appendix~\ref{app:mt_parameters}.
%Each translation and its source text formed a translation unit for evaluation.

%%%%%%%%%%%%%%%%%%%%%%%%%%%%%%%%%%%%%%%%%%%%%%%%%%
\subsection{Human Annotations}

%Crowdworks\footnote{\url{https://crowdworks.jp/}}.
Six professional annotators proficient in Japanese and English and another six proficient in both Chinese and English were recruited through crowdsourcing. 
The annotators identified the translation errors and assigned the severity levels based on MQM-Chat specifications using Label Studio\footnote{\url{https://labelstud.io/}}~\cite{LabelStudio}.
The annotators were provided with detailed guidelines to ensure sufficient consistency in the error labeling and severity assessment consistency.
In addition, we manually reviewed the human annotation results and made necessary corrections to ensure the quality of the annotations.
The reviewer examined the annotations, primarily focusing on whether the label and the severity of the annotations matched their definitions and whether the error span was overly broad.
For example, if there was only a single error word but the error span contained the entire sentence, the reviewer would correct the span.
Two reviewers familiar with MQM-Chat error types and proficient in Chinese, Japanese, and English checked the annotations afterward.
Details about annotation tasks are presented in Appendix~\ref{app:annotations}.

To ensure the quality and safety of the dataset, the annotators were empowered to report any data containing extremely offensive or toxic content, which was implemented to avoid including highly toxic data in the experiments, thereby ensuring a more ethical and controlled evaluation process.
The reported data were reviewed and excluded if deemed inappropriate to maintain the integrity and safety of the annotating task environment.

%%%%%%%%%%%%%%%%%%%%%%%%%%%%%%%%%%%%%%%%%%%%%%%%%%
%\subsection{Metrics Evaluation}

%In this experiment, we assess the effectiveness of MQM-Chat as an evaluation metric.
%\paragraph{Overall Quality Score}
%We derive the Overall Quality Score (OQS) from the standard MQM to assess the performance of the translation models, as shown in Equation~\ref{eq:oqs} where the OQS states Overall Quality Scores, EC states error count, SPM states severity penalty multiplier, EWC states Evaluation Word Count (EWC), and j states the severity level index.
%ここで \small を使うのはまずいので，normal sizeで書きましょう．
%入らなければ，省略形を使って式の前か後ろで省略形を説明しましょう．
%\small
%\begin{equation}
%    OQS = 1 - \frac{\Sigma_j EC_j * SPM_j}{EWC}
%\label{eq:oqs}
%\end{equation}
%\normalsize 
%Three levels of severity and corresponding severity penalty multipliers, 5 for major, 3 for minor, and 1 for neutral, are provided for OQS calculations.
%Based on the OQS, we calculated the system-level pairwise accuracy and Pearson agreements between MQM-Chat and standard MQM.
%\paragraph{Error Type Distribution}
%We analyze the distribution of error types to evaluate the performance of standard MQM and MQM-Chat on the dataset and the translation models used in this study.
%\paragraph{Comparison of Error Types}
%To highlight the differences, we examine errors categorized under standard MQM but classified as chat-specific error types in MQM-Chat.

%%%%%%%%%%%%%%%%%%%%%%%%%%%%%%%%%%%%%%%%%%%%%%%%%%%%%%%%%%%%%%%%%%%%%%%%%%%%%%%%%%%%%%%%%%%%%%%%%%%%
\section{Results and Analysis}

%%%%%%%%%%%%%%%%%%%%%%%%%%%%%%%%%%%%%%%%%%%%%%%%%%
\subsection{Error Distributions of MQM-Chat}

\begin{figure*}[ht]
\small
\centering
\includegraphics[width=\textwidth]{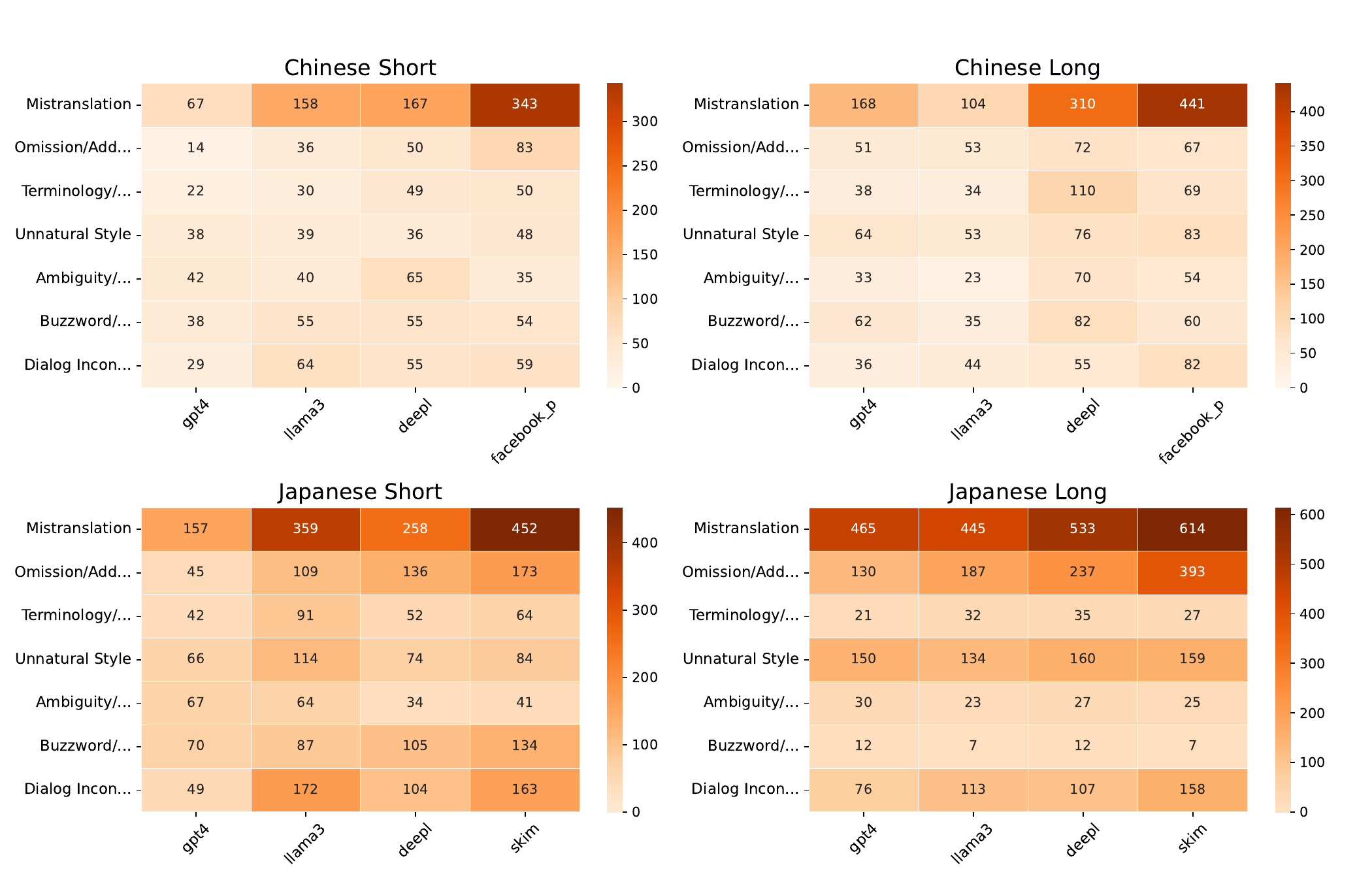}
%(a) MQM-Chat
%\includegraphics[width=\textwidth]{figure/error_counts_heatmap.human-core.pdf}
%(b) Standard MQM (sampled)
\caption{Heatmaps of the error numbers in MQM-Chat %and standard MQM 
human annotations.
%Standard MQM human annotations are applied to 1/4 data.
Darker colors indicate higher numbers. %Error types that are not labeled in any of the four model translations are not shown.
}
\label{fig:error_counts_heatmap}
\end{figure*}

As shown in Figure~\ref{fig:error_counts_heatmap}, we analyzed the error distributions of MQM-Chat annotations.
The results demonstrate that, although the distribution of \textit{Mistranslations} was skewed, MQM-Chat provided a varied distribution of errors across other categories.
A possible reason for having more mistranslations could be that not all the translation models were specifically trained or fine-tuned on parallel zh-en or ja-en chat translation data.

When chat-specific errors occurred, MQM-Chat provided several insights.
First, the ja$\Rightarrow$en translations generally exhibited more errors than the zh$\Rightarrow$en translations.
In terms of the chat length, \textit{Mistranslations} and \textit{Unnatural Style} errors tended to occur frequently in the long chat translations compared to the short chats in both language pairs. 
Furthermore, \textit{Omissions and Additions} were common in long chats, especially in the ja$\Rightarrow$en translations. 
In contrast, \textit{Ambiguity} and \textit{Buzzword Issues} appeared in short chats more frequently.
In addition, \textit{Dialogue Inconsistency} issues were found to be persistent across both the short and long chats, regardless of the source language.

For the ja$\Rightarrow$en translations, regardless of the chat length, the most frequent errors included \textit{Omissions or Additions}, \textit{Unnatural Style}, and \textit{Dialogue Inconsistency} issues.
We found that \textit{Ambiguity and Disambiguation} errors were less common in the ja$\Rightarrow$en translations; however, they occurred at a similar frequency to other errors in the zh$\Rightarrow$en translations.
This could be because Chinese people frequently omit punctuation when chatting, thereby leading to a more even distribution of ambiguous content in the Chinese segments.
In addition, \textit{Terminology and Proper Noun Issues}, \textit{Ambiguity and Disambiguation}, and \textit{Buzzword or Loanword} errors occurred more frequently in the short Japanese chats than in the long chats.
Overall, the findings demonstrate that MQM-Chat provides valuable insights into the overall trends across languages and offers a deeper understanding of the translation challenges associated with different source languages. %which standard MQM cannot clearly identify.

In terms of the translation models, GPT-4 generated considerably fewer \textit{Mistranslations} in the short chat translation tasks than the other three models.
However, the amounts were similar for the long chat translation tasks.
GPT-4 produced more \textit{Buzzword or Loanword Issues} in long chats, and Llama3 struggled with dialogue consistency in the short chats and produced more \textit{Terminology or Proper Noun Issues} and \textit{Unnatural Style} problems in the short Japanese chat translation tasks.
Furthermore, DeepL produced more \textit{Terminology and Proper Noun Issues} in the zh$\Rightarrow$en translations.
The results of MQM-Chat help us better understand the strengths and weaknesses of different models, thereby offering pathways to improve such models.
For example, leveraging the glossary function in DeepL could help reduce terminology errors.

%%%%%%%%%%%%%%%%%%%%%%%%%%%%%%%%%%%%%%%%%%%%%%%%%%
\subsection{Errors Relabeled by MQM-Chat}

\begin{table*}[ht]
\centering
\small
\tabcolsep=3pt
\begin{tabular}{@{}llccccccc@{}}
\toprule
&&&
\multicolumn{2}{c}{\textit{\textbf{zh$\Rightarrow$en}}} && \multicolumn{2}{c}{\textit{\textbf{ja$\Rightarrow$en}}} \\ 
 \textbf{Data} & \textbf{Model} && \textbf{Relabeled (\%)} & \textbf{Chat-spec (\%)} && \textbf{Relabeled (\%)} & \textbf{Chat-spec (\%)} \\
\cmidrule{1-2}
\cmidrule{4-5}
\cmidrule{7-8}
Short 
&\textbf{GPT-4} && 39 (50.65\%) & 23 (29.87\%) && 92 (57.86\%) & 41 (25.79\%) \\
&\textbf{LLaMA3} && 46 (48.94\%) & 29 (30.85\%) && 201 (74.17\%) & 80 (29.52\%) \\
&\textbf{DeepL} && 46 (52.87\%) & 18 (20.69\%) && 172 (81.13\%) & 77 (36.32\%) \\
&\textbf{NMT} && 94 (67.63\%) & 35 (25.18\%) && 329 (86.58\%) & 120 (31.58\%) \\
\cmidrule{1-2}
\cmidrule{4-5}
\cmidrule{7-8}
Long
&\textbf{GPT-4} && 16 (27.59\%) & 3 (5.17\%) && 33 (25.19\%) &10 (7.63\%) \\
&\textbf{LLaMA3} && 21 (26.25\%) & 10 (12.50\%) && 69 (30.53\%) & 17 (7.52\%) \\
&\textbf{DeepL} && 71 (49.65\%) & 34 (23.78\%) && 92 (43.81\%) & 26 (12.38\%) \\
&\textbf{NMT} && 128 (55.41\%) & 48 (20.78\%) && 278 (61.50\%) & 49 (10.84\%) \\ \bottomrule
\end{tabular}
\caption{Percentage of errors labeled with standard MQM human annotations were relabeled with MQM-Chat human annotations. \textbf{Relabeled (\%)} represents the number and percentage of errors being re-labeled as other error types in MQM-Chat human annotations. \textbf{Chat-spec (\%)} indicates the percentage of errors relabeled to chat-specific error types, such as \textit{Ambiguity}, \textit{Buzzword Issues}, and \textit{Dialogue Inconsistency}.}
\label{tab:mistranslation_relabeled_comparison}
\end{table*}

\begin{table*}[ht]
\centering
\small
\begin{tabular}{lp{12cm}}
\toprule

\multicolumn{2}{l}{Example 1} \\ \midrule
source & \zh{我在}\textcolor{blue}{\zh{那滴 (a typographical error of ``那地'')}}\zh{吃的饭。。} \\
NMT output & I had my meals \textcolor{red}{in that drop}.... \\
(possible reference) & \textcolor{blue}{theere} (a typographical error of ``there'') \\
\textbf{Standard MQM} & \textit{Mistranslation} - Critical \\ 
\textbf{MQM-Chat} & \textbf{\textit{Ambiguity and Disambiguation} - Major} \\
\midrule

\multicolumn{2}{l}{Example 2} \\ \midrule
source & \zh{...我只是为了凹造型}\textcolor{blue}{\zh{人艰勿拆}} \\
DeepL output & I just for the sake of the shape \textcolor{red}{of the people hard not to break down} \\
(possible reference) & \textcolor{blue}{Ren Jian Wu Chai} (Chinese transliteration) or \textcolor{blue}{life is already so hard or arduous, so don't judge me.} (the meaning) \\
\textbf{Standard MQM} & \textit{Mistranslation} - Major \\ 
\textbf{MQM-Chat} & \textbf{\textit{Buzzword or Loanword Issues} - Major} \\
\midrule

\multicolumn{2}{l}{Example 3} \\ \midrule
\multirow{2}{*}{source} & \ja{...結婚して早く家を出ろって母がうるさくて。} \\
 & \ja{- そうだったのかぁ。}\textcolor{blue}{\ja{うち(*1)}}\ja{とは逆だね。}\textcolor{blue}{\ja{うちは一緒にいてほしいみたいだよ。(*2)}} \\
\multirow{2}{*}{NMT output} & ...my mother insisted that I get married and leave the house as soon as possible. \\
 & - I see, it's the opposite of \textcolor{red}{my house(*1)}. \textcolor{red}{I want you to stay with me.(*2)} \\
(possible reference) & \textcolor{blue}{(*1) my family, (*2) She want me to stay with her.} \\
\textbf{Standard MQM} &(*1) \textit{Mistranslation} - Major, (*2) \textit{Mistranslation} - Critical \\
\textbf{MQM-Chat} &(*1) \textit{Mistranslation} - Major, (*2) \textbf{\textit{Dialogue Inconsistency} - Major} \\
\bottomrule
\end{tabular}
\caption{Examples of errors being labeled as \textit{Mistranslations} by standard MQM annotations that were classified into chat-specific labels such as \textit{Ambiguity and Disambiguation}, \textit{Buzzword or Loanword Issues}, and \textit{Dialogue Inconsistency} in MQM-Chat annotations. \textbf{Standard MQM} and \textbf{MQM-Chat} columns show the judgements on the MT output by each annotation criteria ([label] - [severity]).}
\label{tab:relabel_examples}
\end{table*}

To demonstrate the superiority of MQM-Chat over existing evaluation metrics in terms of capturing chat-specific translation errors, we compared the behavior of MQM-Chat and standard MQM on the same datasets.
Here, we assigned the standard MQM human annotations, including
\textit{Accuracy} (\textit{Addition}, \textit{Mistranslation}, \textit{Omission}, \textit{Untranslated Text}), \textit{Fluency} (\textit{Character Encoding}, \textit{Grammar}, \textit{Inconsistency}, \textit{Punctuation}, \textit{Register}, \textit{Spelling}), \textit{Locale convention} (\textit{Currency}, \textit{Date}, \textit{Name}, \textit{Telephone}, or \textit{Time Format}), \textit{Style} (\textit{Awkward}), \textit{Terminology} (\textit{Inappropriate for Context}, \textit{Inconsistent Use}), \textit{Non-translation} and \textit{Others}~\cite{10.1162/tacl_a_00437}.
The standard MQM human annotations were applied to 25\% of the data as samples.

To determine the extent of the actual effect of shifting to MQM-Chat, we evaluated the percentage of standard MQM's errors that were labeled as other types by MQM-Chat.
The results are shown in Table~\ref{tab:mistranslation_relabeled_comparison}.
The results suggest that MQM-Chat can successfully recognized at least 25.19\% and at most 86.58\% of the errors.
Half of those errors were labeled as chat-specific errors with the help of MQM-Chat.
Especially for the short chat, the percentage of errors labeled as chat-specific types was higher than that for the long chat, which also meets the nature that short data from LCCC-base and Open 2ch Dialogue Corpus include more source-side ambiguity, buzzword, and inconsistency issues.

Three examples are listed according to the three chat-specific error types in Table~\ref{tab:relabel_examples}. 
In the \textit{Ambiguity and Disambiguation} example, misspelling \zh{那地} as \zh{那滴} caused the translation model to generate an error, which was labeled as \textit{Ambiguity and Disambiguation} according to MQM-Chat, while standard MQM marked this error as a \textit{Mistranslation}.
The second example shows an error caused by the Internet slang \zh{人艰勿拆}, which was labeled as a \textit{Buzzword Issue} by MQM-Chat but a \textit{Mistranslation} by standard MQM.
In the third example, MQM-Chat labeled a sentence with referential problems as \textit{Dialogue Inconsistency} rather than \textit{Mistranslation}.
%We consider that standard MQM labeled the sentence as \textit{Mistranslation} because there is no specific error type to indicate the referential problem happening in conversations.
%The inconsistency-related labels are inconsistent style in standard MQM, not a mistranslation caused by subject inconsistency.

%\input{table/system_level_pairwise_pearson}

%We calculated the system-level accuracy of MQM-Chat human annotations and standard MQM automatic annotations based on OQS to assess the consistency of their evaluations across all systems.%, as shown in Table~\ref{tab:system_level_pairwise_pearson}.
%Based on translation models and datasets, there are 16 systems included in the calculations.
%From the pairwise accuracy, we can conclude that MQM-Chat and standard MQM agree with each other about the quality of translation models since they have a 75\% pairwise accuracy when ranking the translation systems.
%The Pearson scores also suggest a strong agreement between the two metrics.

%In conclusion, based on the OQS, MQM-Chat can have relatively consistent evaluation results with standard MQM.
From the comparison with standard MQM, we believe that MQM-Chat differs from existing evaluation benchmarks because it can locate and analyze problems specific to chat translation precisely.

%%%%%%%%%%%%%%%%%%%%%%%%%%%%%%%%%%%%%%%%%%%%%%%%%%%%%%%%%%%%%%%%%%%%%%%%%%%%%%%%%%%%%%%%%%%%%%%%%%%%
\section{Automatic MQM-Chat Annotations}

\begin{figure*}[ht]
\small
\centering
\includegraphics[width=\textwidth]{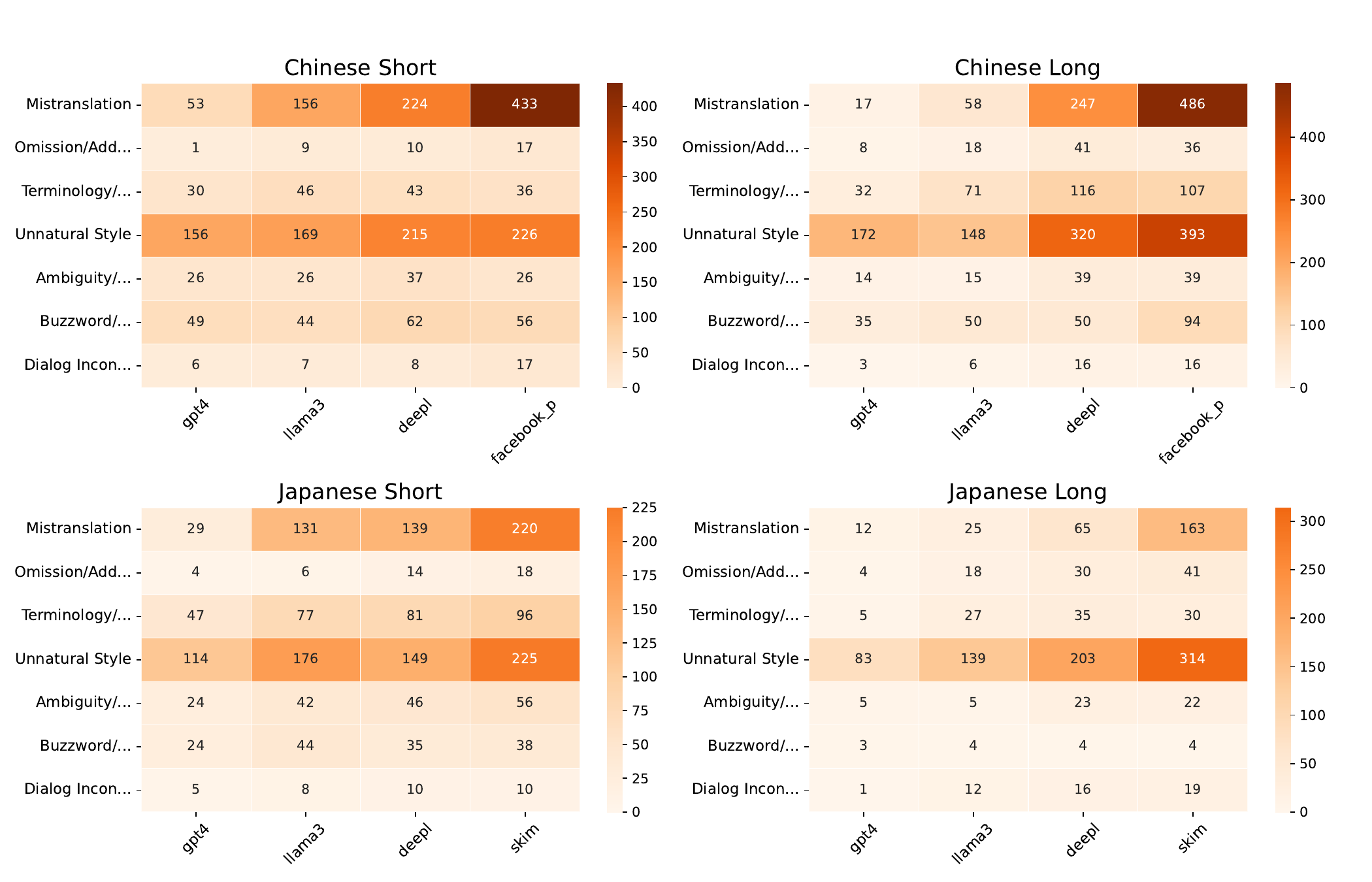}
\caption{Heatmaps of error numbers in MQM-Chat auto annotations.
Darker colors indicate higher numbers.}
\label{fig:error_counts_heatmap_auto}
\end{figure*}

% \usepackage{booktabs}
\begin{table}[t]
\centering
\tabcolsep=3pt
\small
%\begin{tabular}{@{}cccccccccc@{}}
%\toprule
%\multicolumn{10}{c}{\textit{Chinese-to-English}} \\ \midrule
% & \multicolumn{3}{c}{\textbf{Span}} & \multicolumn{3}{c}{\textbf{Span-Label}} & \multicolumn{3}{c}{\textbf{Span-Severity}} \\
% & \textbf{Pre} & \textbf{Rec} & \textbf{F1} & \textbf{Pre} & \textbf{Rec} & \textbf{F1} & \textbf{Pre} & \textbf{Rec} & \textbf{F1} \\
%\multicolumn{1}{l}{\textbf{Short}} & 52.24 & 64.99 & 54.03 & 24.26 & 27.47 & 24.38 & 31.35 & 37.54 & 32.11 \\
%\multicolumn{1}{l}{\textbf{Long}} & 33.97 & 43.56 & 33.45 & 14.11 & 16.54 & 13.55 & 16.69 & 20.64 & 16.33 \\ \midrule
%% & \multicolumn{3}{c}{\textbf{Span}} & \multicolumn{3}{c}{\textbf{Span-Label}} & \multicolumn{3}{c}{\textbf{Span-Severity}} \\
% & \textbf{Pre} & \textbf{Rec} & \textbf{F1} & \textbf{Pre} & \textbf{Rec} & \textbf{F1} & \textbf{Pre} & \textbf{Rec} & \textbf{F1} \\
%\multicolumn{1}{l}{\textbf{Short}} & 55.93 & 42.26 & 43.65 & 20.28 & 15.81 & 16.28 & 29.67 & 21.97 & 23.05 \\
%\multicolumn{1}{l}{\textbf{Long}} & 38.74 & 18.63 & 22.30 & 10.75 & 5.45 & 6.49 & 17.69 & 8.82 & 10.39 \\ \bottomrule
%\end{tabular}
\begin{tabular}{@{}lccccccc@{}}
\toprule
 & \multicolumn{3}{c}{\textit{zh$\Rightarrow$en}} && \multicolumn{3}{c}{\textit{ja$\Rightarrow$en}} \\ \midrule

\multicolumn{1}{c}{} & \multicolumn{3}{c}{\textbf{Span}} && \multicolumn{3}{c}{\textbf{Span}} \\
\textbf{Data} & \textbf{Pre} & \textbf{Rec} & \textbf{F1} && \textbf{Pre} & \textbf{Rec} & \textbf{F1} \\
\cmidrule{2-4}
\cmidrule{6-8}
Short & 52.24 & 64.99 & 54.03 && 55.93 & 42.26 & 43.65 \\
Long & 33.97 & 43.56 & 33.45 && 38.74 & 18.63 & 22.30 \\ \midrule

\multicolumn{1}{c}{\textit{}} & \multicolumn{3}{c}{\textbf{Span+Label}} && \multicolumn{3}{c}{\textbf{Span+Label}} \\
\textbf{Data} & \textbf{Pre} & \textbf{Rec} & \textbf{F1} && \textbf{Pre} & \textbf{Rec} & \textbf{F1} \\
\cmidrule{2-4}
\cmidrule{6-8}
Short & 24.26 & 27.47 & 24.38 && 20.28 & 15.81 & 16.28 \\
Long & 14.11 & 16.54 & 13.55 && 10.75 & 5.45 & 6.49 \\ \midrule

\multicolumn{1}{c}{} & \multicolumn{3}{c}{\textbf{Span+Severity}} && \multicolumn{3}{c}{\textbf{Span+Severity}} \\
\textbf{Data} & \textbf{Pre} & \textbf{Rec} & \textbf{F1} && \textbf{Pre} & \textbf{Rec} & \textbf{F1} \\
\cmidrule{2-4}
\cmidrule{6-8}
Short & 31.35 & 37.54 & 32.11 && 29.67 & 21.97 & 23.05 \\
Long & 16.69 & 20.64 & 16.33 && 17.69 & 8.82 & 10.39 \\ 
\bottomrule
\end{tabular}
\caption{Average precision, recall and F1 scores of MQM-Chat automatic annotations having span overlap, span and label overlap, span and severity overlap with human annotations.}
\label{tab:span_overlap_average}
\end{table}

%We conducted the experiment of automatic MQM-Chat evaluation through few-shot learning on GPT-4.
We implemented MQM-Chat automatic evaluation based on the GEMBA-MQM~\cite{kocmi-federmann-2023-gemba} prompt by replacing the description of the error types of standard MQM with the error types of MQM-Chat.
Three chat translations with MQM-Chat annotations were provided as examples for few-shot learning.
The prompt is shown in Appendix~\ref{app:prompt_and_examples}.
Using human annotations as the golden standard, we calculated the pairwise accuracy and Pearson agreements for the automatic evaluation. %, as shown in Table~\ref{tab:system_level_pairwise_pearson}.
The 79.17\% pairwise accuracy and 0.774 Pearson correlation values demonstrate that the auto annotations of MQM-Chat agree with human rankings.

We also reviewed the error spans of the proposed MQM-Chat to investigate whether its annotations can also analyze errors in detail.
The distribution of the error numbers is shown in Figure~\ref{fig:error_counts_heatmap_auto}.
Unlike human annotations, automatic annotations focus on \textit{Mistranslation} and \textit{Unnatural Style} in all four cases.
The number of \textit{Unnatural Style} errors was even greater than that of \textit{Mistranslations} for the LLMs' zh$\Rightarrow$en translations.
Additionally, fewer errors were observed in ja$\Rightarrow$en than in zh$\Rightarrow$en, with always more \textit{Unnatural Style} errors, which is the opposite of the human annotations,
The findings may be related to the limited amount of Japanese chat data, compared with the Chinese chat data, in GPT's training dataset.

Span-level accuracy is based on the methodology employed in AutoMQM; however, a more flexible approach is adopted by considering spans as overlapping if they mostly align rather than strictly counting overlapping characters.
Here, we calculated precision, recall, and F1 scores for three scenarios: span overlap, both span and label overlap, and both span and severity overlap, as shown in Table~\ref{tab:span_overlap_average}.
The results demonstrate that MQM-Chat auto annotations have limited overlap with human annotations, particularly with the long chat translation tasks.
We consider that the current few-shot prompting approach with GPT-4 is insufficient to fully support the automatic evaluation of MQM-Chat.
However, this is understandable, given that MQM-Chat is a new evaluation metric.
We plan to optimize MQM-Chat automatic annotations by selecting open-source LLMs and utilizing human annotations for fine-tuning in the future.

%%%%%%%%%%%%%%%%%%%%%%%%%%%%%%%%%%%%%%%%%%%%%%%%%%%%%%%%%%%%%%%%%%%%%%%%%%%%%%%%%%%%%%%%%%%%%%%%%%%%
\section{Conclusion}

To address the lack of evaluation metrics for chat translation tasks, this study has proposed MQM-Chat and evaluated its effectiveness through a series of experiments.
By analyzing the error distribution, we find that MQM-Chat is suitable for qualifying chat translations and can successfully identify the weakness of the experimented translations.
Looking at the errors that were relabeled as other error types by MQM-Chat, we consider that MQM-Chat can provide a more nuanced classification of errors, especially for chat-specific issues.

In addition, we explored MQM-Chat automatic annotations by few-shots learning on GPT-4 with GEMBA-MQM's prompt.
It agrees with the system rankings annotated by human annotators but cannot fully overlap with the spans in human annotations.
However, although it currently needs further refinements, we still consider that it can serve as a basic reference.

In this study, to address the complexities of translating everyday chat conversations, we selected data that were rich in slang and ambiguous contents, which inherently increases the difficulty of the translation tasks.
In the future, we plan to evaluate MQM-Chat on chat translation data from other domains, such as custom service, to determine if the results have the same characteristics.
In addition, we plan to enhance the implementation of automatic MQM-Chat, thereby enabling it to serve as an effective evaluation reference for chat translation tasks.
Ultimately, we hope that the MQM-Chat can be used as a valuable evaluation benchmark for chat translation tasks to facilitate good performance in translation tasks involving chats and other informal content translations.

\clearpage
%%%%%%%%%%%%%%%%%%%%%%%%%%%%%%%%%%%%%%%%%%%%%%%%%%%%%%%%%%%%%%%%%%%%%%%%%%%%%%%%%%%%%%%%%%%%%%%%%%%%
\section*{Limitations}

With data limited to translations from Chinese and Japanese to English, the experimental results obtained in the current study are relatively narrow.
Thus, in the future, we plan to extend MQM-Chat to more language pairs and bidirectional translations to better understand chat translation across various languages.
%The high frequency of mistranslation errors in our results indicates that this error type needs further refinement.
%We plan to conduct more detailed reviews of the annotations to identify if additional nodes of mistranslation are necessary.
%Additionally, we consider introducing automatic annotation by models to check if they can perform annotation tasks based on existing data and MQM-Chat definitions.
In summary, we consider that MQM-Chat has laid a solid foundation for this type of research, opening up many potential directions to improve and expand chat translation evaluations.

%%%%%%%%%%%%%%%%%%%%%%%%%%%%%%%%%%%%%%%%%%%%%%%%%%%%%%%%%%%%%%%%%%%%%%%%%%%%%%%%%%%%%%%%%%%%%%%%%%%%
\section*{Ethical Considerations}

The crowdsourcing experiments conducted in this study adhered to stringent ethical guidelines to ensure participant privacy and data protection.
The experiments deliberately avoided collecting any personally identifiable information from the participants.
No restrictions or enforcement of specific work hours were imposed upon the participants, thereby eliminating undue influence or coercion.
Given the absence of personal data collection and voluntary participation, the data were not subject to an ethics review at the organization.
Consequently, the data collection procedures used in this study adhered to the ethical standards and regulations governing acceptable research practices.

%%%%%%%%%%%%%%%%%%%%%%%%%%%%%%%%%%%%%%%%%%%%%%%%%%%%%%%%%%%%%%%%%%%%%%%%%%%%%%%%%%%%%%%%%%%%%%%%%%%%
\section*{Acknowledgements}
This work was supported by 
JST (the establishment of university fellowships towards the creation of science technology innovation) Grant Number JPMJFS2102, % 博士フェローシップ（～２０２４・４）
JST SPRING Grant Number JPMJSP2114, % 挑戦的プロジェクト（２０２４・４～）
JSPS KAKENHI 22H00524, % 乾さん基盤A アノテーション案件に使用
JST CREST Grant Number JPMJCR20D2 % 乾さん CREST
and JST Moonshot R\&D Grant Number JPMJMS2011-35 (fundamental research). % 鈴木さん
The crowdsourcing was supported by Crowdworks (\url{https://crowdworks.jp/}).
The annotation was supported by Label Studio (\url{https://labelstud.io/}), especially thanks for providing the academic version.

\bibliography{anthology, mycitaion, custom}

%%%%%%%%%%%%%%%%%%%%%%%%%%%%%%%%%%%%%%%%%%%%%%%%%%%%%%%%%%%%%%%%%%%%%%%%%%%%%%%%%%%%%%%%%%%%%%%%%%%%
\clearpage

\appendix
%%%%%%%%%%%%%%%%%%%%%%%%%%%%%%%%%%%%%%%%%%%%%%%%%%
\section{Datasets and Licenses}
\label{app:datasets}

\begin{table*}[t]
\centering
\begin{tabular}{@{}lcccc@{}}
\toprule
 & \textbf{LCCC-base} & \textbf{NaturalConv} & \textbf{Open2ch Dialogue} & \textbf{BPersona-chat} \\ \midrule
\textbf{Source Language}                  & Chinese & Chinese & Japanese & Japanese \\
\textbf{\# of chats selected}      & 200     & 100     & 200      & 100      \\
\textbf{Avg.turns}                 & 5       & 21      & 5        & 12       \\
\textbf{Avg. \# of char (src)}     & 52      & 423     & 101      & 490      \\
\textbf{Avg. \# of words (GPT-4)}  & 38      & 248     & 52       & 214      \\
\textbf{Avg. \# of words (LLaMA3)} & 37      & 235     & 50       & 203      \\
\textbf{Avg. \# of words (DeepL)}  & 36      & 247     & 49       & 218      \\
\textbf{Avg. \# of words (NMT)}    & 37      & 272     & 45       & 182      \\ \bottomrule
\end{tabular}%
\caption{The number of average turns, average source words, average worse in translations of the selected data.}
\label{tab:data_stat}
\end{table*}

The statistical information of the selected monolingual data and their translations are shown in Table~\ref{tab:data_stat}.
The NLTK package~\cite{bird-loper-2004-nltk} was used to calculate the word counts.

All four datasets used in this research come with a license allowing non-commercial and academic usage. To be specific, the licenses are:  
\href{https://github.com/thu-coai/CDial-GPT?tab=MIT-1-ov-file#readme}{MIT License} for LCCC-base~\cite{wang2020chinese}; 
\href{https://ai.tencent.com/ailab/nlp/dialogue/papers/NaturalConv_Release_license.pdf}{Tecent AI Lab NaturalConv Dataset Terms and Conditions} for NaturalConv~\cite{wang2021naturalconv}; 
CC BY-SA 4.0 for Open2ch Dialogue Corpus~\cite{open2chdlc2019en}; 
and CC BY-NC 4.0 for BPersona-chat~\cite{sugiyama-etal-2021-empirical, li-etal-2022-chat}.
The annotated data of this research will also be published in CC BY-NC 4.0 for non-commercial usage in the future.

%%%%%%%%%%%%%%%%%%%%%%%%%%%%%%%%%%%%%%%%%%%%%%%%%%
\section{Machine Translation Parameters}
\label{app:mt_parameters}

\subsection{GPT-4 and LLaMA3}

The prompts used in GPT-4 and LLaMA3's requests were structured as follows:

\begin{lstlisting}
You are a professional {source_language} to 
{target_language} translator. This is a 
{source_language} to {target_language} chat 
translation task. Please translate each line 
of the chat from {source_language} to 
{target_language}. Each line of the chat is 
considered a message sent by a different speaker.
\end{lstlisting}
% Return directly only the {target_language} translation as lines.
Models were set to gpt-4 and Meta-Llama-3-70B-Instruct, respectively.
Other parameters were set to \verb|temperature=1, top_p=1.0, max_token=500|, and defaults.

\subsection{Facebook@WMT21}

The neutral machine translation model used for Chinese to English translations was the multilingual model submitted to WMT 2023 by Facebook~\cite{tran-etal-2021-facebook, akhbardeh-etal-2021-findings}.
The model can directly translate text from 7 languages: Hausa (ha), Icelandic (is), Japanese (ja), Czech (cs), Russian (ru), Chinese (zh), German (de) to English.
For Chinese to English, it was trained on 166M bitext data from the WMT shared task, and 123M monolingual data from Commoncrawl\footnote{\url{http://data.statmt.org/cc-100/}} which are news-domain.
The model consists of a 24-layer encoder/decoder with an embedding size of 2,048 and a feedforward size of 16,384 and 32 attention heads, resulting in 4.7B total parameters.
Trainings were taken on 32 Volta 32GB GPUs.
Fore more details, please refer to the original paper~\cite{tran-etal-2021-facebook}.

\subsection{SKIM@WMT23}
We used a neural machine translation system submitted to WMT 2023 by team SKIM~\cite{kudo-etal-2023-skim}, who achieved the best accuracy among the participants in WMT23~\cite{kocmi-etal-2023-findings}.
The model was trained on publicly available Japanese-English parallel data of around 31M sentences and a synthetic parallel corpus of 681M sentences.
The model consists of a 9-layer encoder/decoder with an embedding size of 1,024 and a feedforward size of 8,192, and 16 attention heads, resulting in 547M total parameters.
Training took around four days with eight NVIDIA RTX A6000 GPUs.
For more details of training settings, please refer to the original paper~\cite{kudo-etal-2023-skim}.

%%%%%%%%%%%%%%%%%%%%%%%%%%%%%%%%%%%%%%%%%%%%%%%%%%
\section{Crowdsourcing Annotation Tasks}
\label{app:annotations}

\subsection{Crowdsourcing Annotators}

Considering that chat translation requires not only proficiency in two languages but also an understanding of the source text, we called for native Chinese or Japanese speakers who are fluent in English to be the annotators through crowdsourcing platforms.
%We require candidates to provide an English major degree, test scores, or proof of overseas residency to ensure their English language proficiency.
We prepared qualifications for the candidates to determine their suitability for the task, which consisted of five short chats and three long chats. % with reference annotations.
Participants who showed a better understanding of both the source and target languages were considered to meet our expectations better and were selected as annotators.
All annotators are aware that their annotations will be used for academic research, not commercial.

\subsection{Annotating Instructions}

Annotators were provided with detailed instructions in English, Chinese, and Japanese.
The instructions include the labeling descriptions with Label Studio and the definitions of error types and severities.
Each error type and severity level was provided with 1-5 detailed examples to help annotators understand.
Annotators are instructed and required to report offensive data when the source contains extremely offensive content as well.
The reported data are removed to avoid having toxic contents in the annotated dataset.

\subsection{Reviewing Instructions}

Reviewers were provided with the annotating instructions and the additional reviewing instructions.
The reviewing instructions introduce rules for reviewing and correcting annotations, such as: 
\begin{itemize}
    \item if the error happens on a single word, remove punctuations before or after.
    \item if the error only has punctuations, keep the punctuations.
    \item if there are multiple labels, check each label and remove some of them if needed, especially when having mistranslations with other labels together.
    \item dialogue inconsistency errors should be treated separately; if the inconsistency problem is the same (i.e., `you' and `yours' in one sentence), the second could be neutral, otherwise, the second should be labeled as major or minor.
    \item if a buzzword is missed, label the words or punctuation before the losing buzzword with buzzword and omission together.
\end{itemize}
Other rules are present as well to refine the annotations.

\subsection{Payments}
We paid each annotator an extra 30-35 USD to familiarize them with the instructions and operations.
Being familiar with the instruction and operation of Label Studio, the annotator took about 3-5 minutes to complete one short chat and about 5-8 minutes for a long chat. 
Depending on the length of data, each annotator was paid about 0.5-1.5 USD per short chat and 0.7-2 USD per long chat, with the final payment fluctuating according to the exchange rate.
In conclusion, every annotator was paid around 18-22 USD per hour.

%%%%%%%%%%%%%%%%%%%%%%%%%%%%%%%%%%%%%%%%%%%%%%%%%%
\section{Automatic MQM-Chat Prompts and Few-shot Examples}
\label{app:prompt_and_examples}

\begin{figure*}[hbt]{
\begin{lstlisting}
(System) You are an annotator for the quality of machine translation. 
Your task is to identify errors and assess the quality of the translation. 

(user) {source_language} source:
```{source_segment}```
{target_language} translation:
```{target_segment}```

Based on the source segment and machine translation surrounded with triple backticks, identify error
types in the translation and classify them. The categories of errors are: 
Mistranslation - fundamental inaccuracies in the translation process, including untranslated source 
segments, incorrect lexical choice or grammar that distorts the meaning, under-translation, and 
over-translation.
Omission or Addition - missing source contents (omission) or additional content not present in the
source (addition).
Terminology or Proper Noun Issues - pertain to inaccuracies in translating specialized vocabulary, 
inherent terms, and proper nouns from the source text.
Unnatural Style - grammatically correct but not natural in the target language.
Ambiguity and Disambiguation - ambiguities or errors in the source text (i.e. typos, graphical 
errors, omissions, unclear abbreviations, erroneous punctuation) are not faithfully reflected in the
translation.
Buzzword or Loanword Issues - when buzzwords or loanwords (i.e. popular sayings, newly created words,
internet slang, memes) are not translated accurately according to their usage in the source and 
target languages.
Dialogue Inconsistency - fail to maintain consistency based on context, particularly when the 
speakers change within the chat.
other, or no-error.
Each error is classified as one of three categories: major, minor, and neutral. Major for errors that 
significantly impact the understandability of the content; 'minor for errors that do not impact the 
overall understandability but detract from the quality; neutral for errors requiring additional 
revision but do not pose significant risks.

(assistant) {observed error classes}
\end{lstlisting}
}
\caption{The prompt for MQM-Chat automatic annotations based on GEMBA-MQM.}
\label{fig:mqmchat_prompt}
\end{figure*}

The MQM-Chat auto annotation experiments used the prompt based on GEMBA-MQM, as shown in Figure~\ref{fig:mqmchat_prompt}.
%The examples used for the few-shot learning are shown in Figure~\ref{fig:few_shots}.

%%%%%%%%%%%%%%%%%%%%%%%%%%%%%%%%%%%%%%%%%%%%%%%%%%
\section{Human Annotation Results}
\label{app:human_annotation_result}

% \usepackage{booktabs}
\begin{table}[t]
\small
\centering
\begin{tabular}{@{}llcccccc@{}}
\toprule
&&& \multicolumn{2}{c}{zh $\Rightarrow$ en} && \multicolumn{2}{c}{ja $\Rightarrow$ en} \\ 
\textbf{Data} &\textbf{Model}&& \textbf{OQS} & \textbf{EC} && \textbf{OQS} & \textbf{EC} \\
\cmidrule{1-2}
\cmidrule{4-5}
\cmidrule{7-8}
\textbf{Short}
&\textbf{GPT-4} && 86.68 & 251 && 82.96 & 498 \\
&\textbf{LLaMA3} && 75.76 & 424 && 59.31 & 999 \\
&\textbf{DeepL} && 73.42 & 477 && 72.16 & 763 \\
&\textbf{NMT} && 53.11 & 672 && 51.31 & 1111 \\
\cmidrule{1-2}
\cmidrule{4-5}
\cmidrule{7-8}
\textbf{Long}
&\textbf{GPT-4} && 94.63 & 453 && 89.63 & 887 \\
&\textbf{LLaMA3} && 95.14 & 346 && 83.77 & 943 \\
&\textbf{DeepL} && 89.03 & 775 && 85.09 & 1112 \\
&\textbf{NMT} && 87.43 & 856 && 73.06 & 1384 \\ \bottomrule
\end{tabular}
\caption{The overall quality score (OQS) and total error counts (EC) of translation models based on human MQM-Chat annotations.}
\label{tab:human_oqs_ec}
\end{table}

Using the MQM-Chat human annotations, we calculated the average OQS and total error count, as shown in Table~\ref{tab:human_oqs_ec}, to compare the overall performance of the translation models.
The OQS and error counts suggest that models perform better on longer chats than shorter ones.
This might be because longer chats generally contain fewer buzzwords and ambiguities, making the translation task more akin to traditional document translation.
The results indicate that zh$\Rightarrow$en translations have higher overall quality and significantly fewer errors than ja$\Rightarrow$en translations.
GPT-4 outperformed the other models in most cases, while the NMT models showed the poorest performance for both language pairs.
LLaMA3 surpassed DeepL in Chinese translations but lagged behind DeepL in Japanese translations, especially in short chats.
This difference could be due to a lack of training data for Japanese and limited language transfer capabilities.

%%%%%%%%%%%%%%%%%%%%%%%%%%%%%%%%%%%%%%%%%%%%%%%%%%
\subsection{Error Distribution in Standard MQM Annotations}

\begin{figure*}[ht]
\small
\centering
\includegraphics[width=\textwidth]{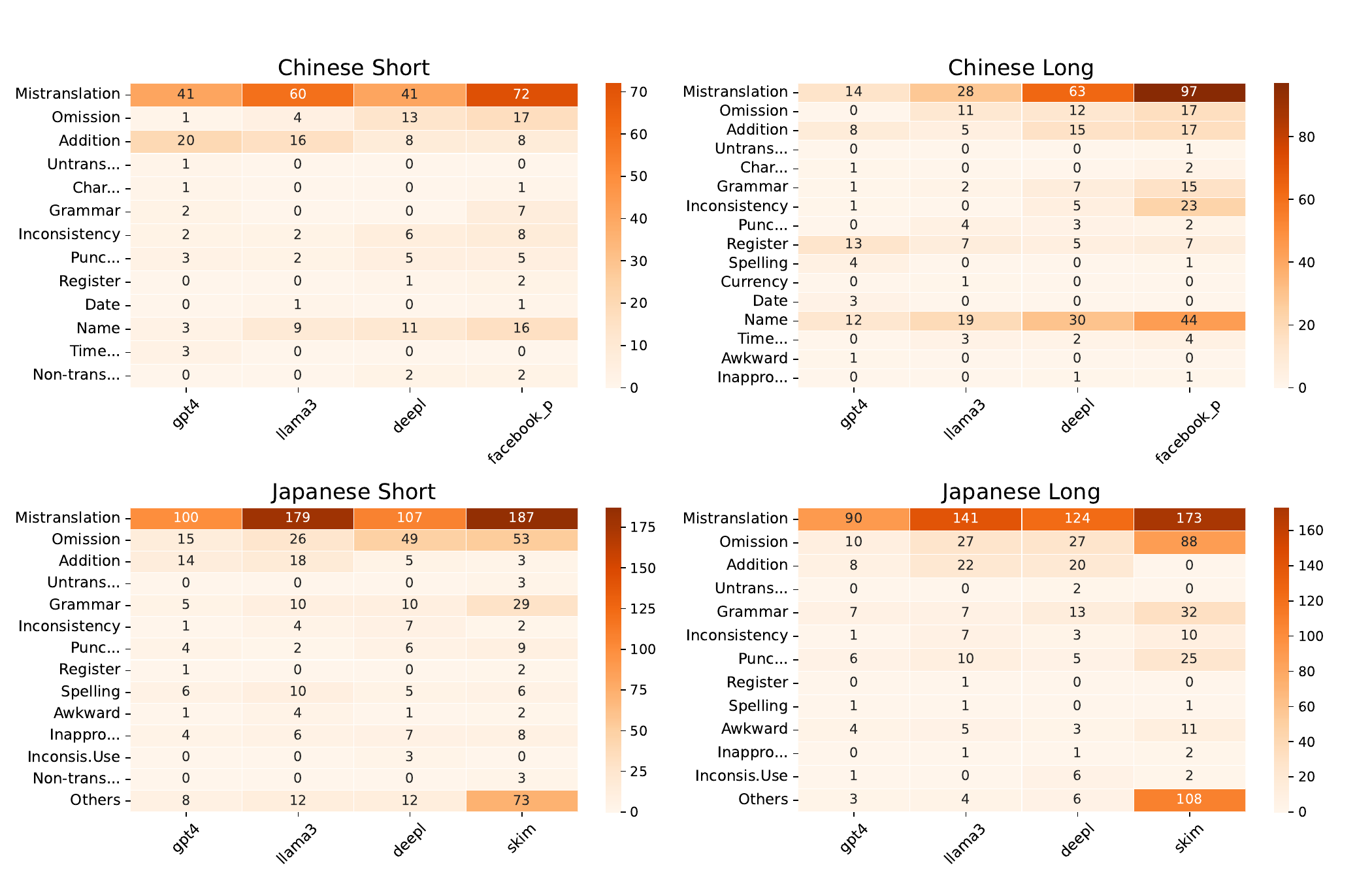}
(a) Human Annotations
\includegraphics[width=\textwidth]{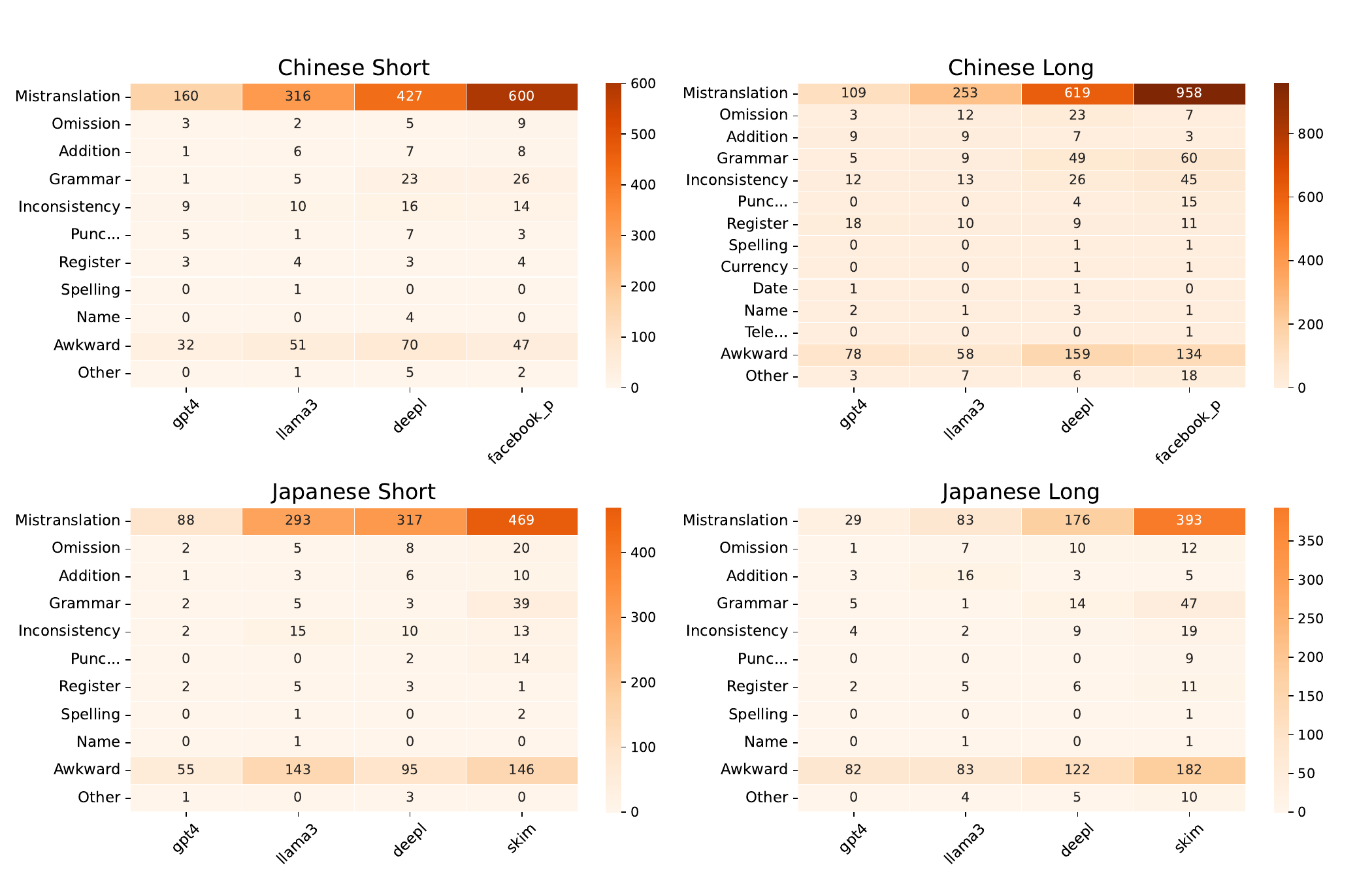}
(b) GPT-4's Annotations
\caption{Heatmaps of the error numbers in standard MQM human and auto annotations.
Darker colors indicate higher numbers.
Standard MQM human annotations are applied to 25\% data.
Error types not labeled in any of the four model translations are not shown.
}
\label{fig:error_counts_heatmap_standard}
\end{figure*}

The distributions of error counts of standard MQM human and automated annotations are shown in Figure~\ref{fig:error_counts_heatmap_standard}.
The standard MQM automated annotations are generated by GPT-4 with GEMBA-MQM's prompt~\cite{kocmi-federmann-2023-gemba}.
The results also show a skewed distribution of Mistranslation, which aligns with the results of MQM-Chat shown in Figure~\ref{fig:error_counts_heatmap} and Figure~\ref{fig:error_counts_heatmap_auto}.
In addition, standard MQM human annotations, even only being tested on 1/4 of the data, show focuses on Omission, Addition, Grammar, and Inconsistency.
The Chinese-to-English translations include Name issues;
meanwhile, the Japanese-to-English translations include Other issues, mainly the whitespace problems.
GPT-4's annotations, instead, show its focus on Awkward Style problems.
The results suggest that standard MQM is not suitable for evaluating chat translations compared to MQM-Chat since it has a less balanced distribution.
\end{document}